\documentclass[11pt]{article}

\usepackage{deauthor}
\usepackage{times}
\usepackage{paralist} 
\usepackage{verbatim}
\usepackage{subcaption}
\usepackage{color}
\usepackage{multirow}
\usepackage{listings}
\usepackage{array}
\usepackage{graphicx}
\usepackage{tabularx}
\usepackage{lscape}
\usepackage{rotating}
\usepackage{algorithm}
\usepackage{algorithmic}
\usepackage{multicol}
\usepackage{xspace}
\usepackage{amsfonts}
\usepackage[noadjust]{cite}
\usepackage{booktabs}

\usepackage{wrapfig}

\usepackage[pdftex]{hyperref}

\def \cD {\mathcal D}
\def \cH {\mathcal H}
\def \cX {\mathcal X}
\def \cY {\mathcal Y}
\newcommand{\RR}{\mathbb{R}}

\begin{document}

\title{A Data Quality-Driven View of MLOps}
\author{Cedric Renggli$^\dagger$~~Luka Rimanic$^\dagger$~~Nezihe Merve Gürel$^\dagger$~~Bojan Karla\v{s}$^\dagger$~~Wentao Wu$^\ddagger$~~Ce Zhang$^\dagger$\\
$^\dagger$ETH Zurich\\
$^\ddagger$Microsoft Research\\
\{cedric.renggli, luka.rimanic, nezihe.guerel, bojan.karlas, ce.zhang\}@inf.ethz.ch\\
wentao.wu@microsoft.com
}

\maketitle

\begin{abstract}
Developing machine learning models can be seen as 
a process similar to the one established for 
traditional software development.
A key difference between the two lies in the 
strong dependency between the quality of a machine learning model
and the quality of the data used to train or perform evaluations.
In this work, we demonstrate how different aspects
of data quality propagate through various stages
of machine learning development.
By performing a joint analysis of the impact 
of well-known data quality dimensions and 
the downstream machine learning process, we show that different components
of a typical MLOps pipeline can be
efficiently designed, providing both a technical and theoretical 
perspective.
\end{abstract}

\section{Introduction}

A machine learning (ML) model 
is a software artifact ``compiled'' from data~\cite{karpathy2017software2}. 
This point of view motivates a study of both similarities
and distinctions when compared to 
traditional software.
\textit{\underline{Similar to}} traditional software artifacts,
an ML model deployed in production 
inevitably undergoes the DevOps process ---
a process whose aim is to
``\textit{shorten the system development life cycle
and provide continuous delivery with 
high software quality}''~\cite{bass2015devops}.
The term ``MLOps'' is used
when this DevOps process is specifically applied to ML~\cite{alla2021mlops}.
\textit{\underline{Different from}}
traditional software artifacts,
the quality of an ML model 
(e.g., accuracy, fairness, and robustness) is often 
a reflection of the
\textit{quality of the underlying data}, 
e.g., noises, imbalances, and additional adversarial perturbations.

Therefore, one of the most promising 
ways to improve the accuracy, fairness, and robustness of an ML model is often to 
improve the dataset, via means such as
data cleaning, integration, and label acquisition. As MLOps aims to 
\textit{understand}, \textit{measure}, and \textit{improve} the quality
of ML models, it is not surprising to see that 
\textit{data quality} is playing a prominent and central role in MLOps. In fact,
many researchers have conducted fascinating and seminal
work around MLOps by looking into different aspects of 
data quality. Substantial effort has been made in the areas of data acquisition with weak supervision (e.g., Snorkel~\cite{ratner2017snorkel}), ML engineering pipelines (e.g., TFX~\cite{katsiapis2019towards}), data cleaning (e.g., ActiveClean~\cite{krishnan2016activeclean}), data quality verification (e.g., Deequ~\cite{schelter2019differential, schelter2018automating}), interaction (e.g., Northstar~\cite{kraska2018northstar}), or fine-grained monitoring and improvement (e.g., Overton~\cite{re2019overton}), to name a few.

Meanwhile, for 
decades data quality has been an active and exciting
research area led by the data management community~\cite{batini2009methodologies,strong1997data,scannapieco2002data},
having in mind that the majority of the studies are 
agnostic to the downstream ML models
(with prominent recent exceptions such as
ActiveClean~\cite{krishnan2016activeclean}).
Independent of downstream ML models,
researchers have studied different aspects of
data quality that can naturally be split across the following four \textit{dimensions}~\cite{batini2009methodologies}: (1)~\textit{accuracy} -- the extent to which the data are correct, reliable and certified for the task at hand;
(2)~\textit{completeness} -- the degree to which the given data collection includes data that describe the corresponding set of real-world objects;
(3)~\textit{consistency} -- the extent of violation of semantic rules defined over a set of data; and 
(4)~\textit{timeliness} (also referred to as \textit{currency} or \textit{volatility}) -- the extent to which data are up-to-date for a task.

\begin{wrapfigure}{r}{0.5\textwidth}
\vspace{-1em}
\begin{center}
\includegraphics[width=0.5\textwidth]{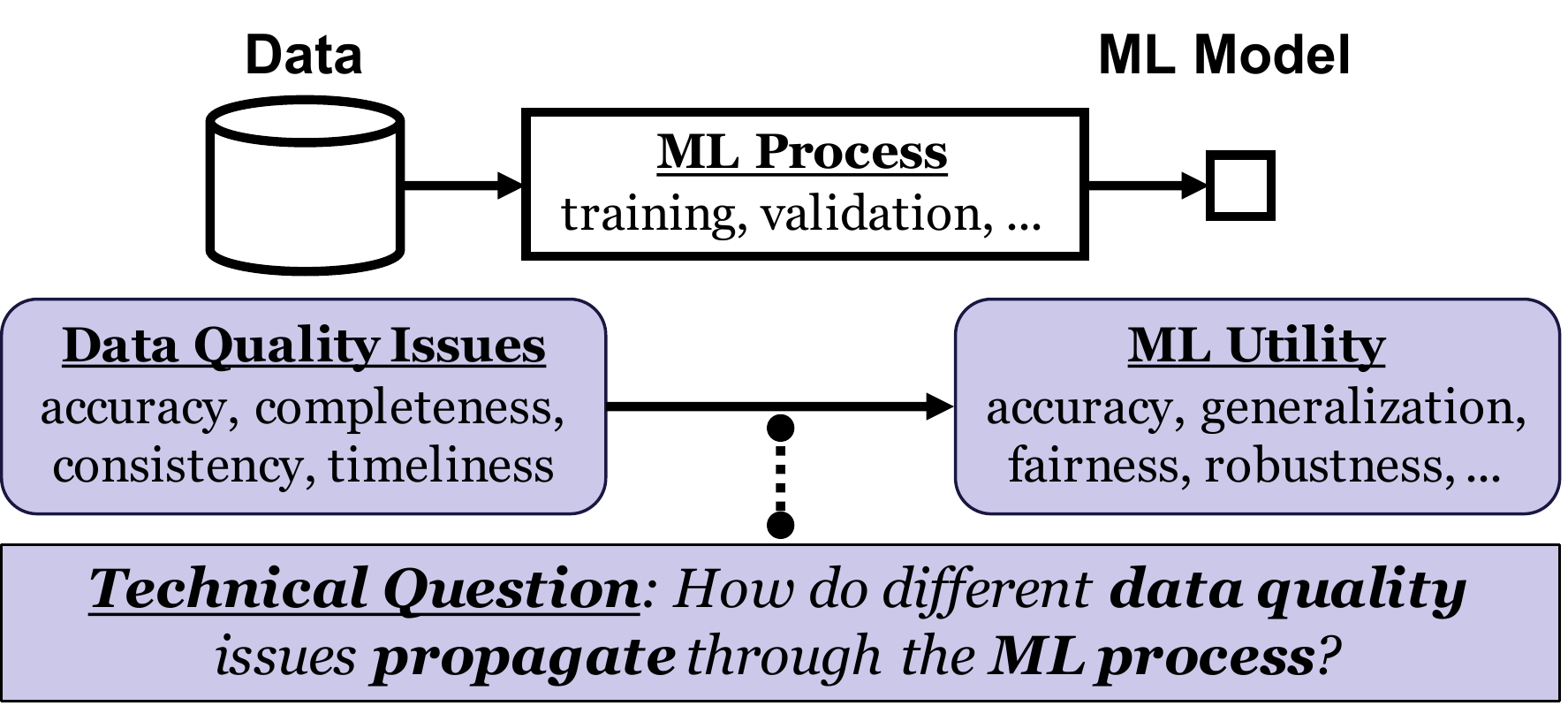}
\end{center}
\vspace{-2em}
\vspace{-1em}
\end{wrapfigure}

\vspace{-0.5em}
\paragraph*{Our Experiences and Opinions} 
In this paper, we 
provide a bird's-eye view 
of some of our previous 
works that are related to
enabling different functionalities with respect to MLOps. These works are 
inspired by our experience working hand-in-hand with academic and industrial users to build
ML applications~\cite{schawinski2017generative, su2018generative, sartori2018model, stark2018psfgan, ackermann2018using, schawinski2018exploring, girardi2018patient, glaser2019radiogan, beck2019sensing, sartori2019forward}, together with  
our effort of building 
\texttt{ease.ml}~\cite{aguilar2021ease}, a prototype
system that defines an 
end-to-end MLOps process.

Our key observation is 
that often \textit{
MLOps challenges are bound to
data management challenges} --- 
given the aforementioned 
strong dependency 
between the quality of 
ML models and the quality of data, the never-ending
pursuit of \textit{understanding,
measuring}, and \textit{improving the
quality of ML models},
often hinges on 
\textit{understanding,
measuring}, and \textit{improving
the underlying 
data quality issues}.
From a technical perspective, this poses 
unique challenges and opportunities.
As we will see, we find it necessary to 
revisit decades of 
data quality research
that are agnostic to 
downstream ML models and 
try to 
understand different 
data quality dimensions -- accuracy, completeness,
consistency, and timeliness -- jointly with the downstream ML process. 

In this paper, we describe 
four of such examples, originated from our 
previous research~\cite{karlavs2020nearest, renggli2020automatic, karimi2020online, renggli2019continuous}.
Table~\ref{tbl:overview}
summarizes 
these examples, each of which tackles one
specific problem in MLOps and
poses technical challenges 
of jointly analyzing data quality 
and downstream ML processes.

\begin{table}[t!]
\caption{Overview of 
our explorations with data quality propagation at different stages of an MLOps 
process.}
\label{tbl:overview}
\vspace{-0.5em}
\small
\centering
\begin{tabular}{@{}llll@{}}
\toprule
\textbf{Technical Problem for ML}                           & \textbf{MLOps Stage}         & \textbf{MLOps Question}                                   & \textbf{Data Quality Dimensions}        \\ \midrule
Data Cleaning (Sec.~\ref{sec:cpclean})~\cite{karlavs2020nearest}              & Pre Training  & Which training sample 
to clean?              &
Accuracy \& Completeness\\
Feasibility Study (Sec.~\ref{sec:snoopy})~\cite{renggli2020automatic}         & Pre Training  & Is my target accuracy realistic? & Accuracy \& Completeness\\
CI/CD (Sec.~\ref{sec:ci})~\cite{renggli2019continuous} & Post Training & 
Am I overfitting to val/test? & Timeliness\\
Model Selection (Sec.~\ref{sec:model_picker})~\cite{karimi2020online}        & Post Training & Which samples should I label?                        & Completeness \& Timeliness\\
\bottomrule
\end{tabular}
\end{table}

\vspace{-0.5em}
\paragraph*{Outline}
In Section~\ref{sec:ml} we provide a setup for studying this topic, highlighting the importance of taking the underlying probability distribution into account. In Sections~\ref{sec:cpclean}-\ref{sec:model_picker} we revisit components of different stages of the \texttt{ease.ml} system purely from a data quality perspective. Due to the nature of this paper, we avoid going into the details of the interactions between these components or
their technical details.
Finally, in Section~\ref{sec:limitations} we describe a common limitation that all the components share, and motivate interesting future work in this area.

\section{Machine Learning Preliminaries}
\label{sec:ml}

In order to highlight the strong dependency between the data samples used to train or validate an ML model and its assumed underlying probability distribution, we start by giving a short primer on ML. In this paper we restrict ourselves on supervised learning in which, given a \textit{feature space} $\cX$ and a \textit{label space} $\cY$, a user is given access to a dataset with $n$ samples  $\cD := \{(x_i, y_i)\}_{i \in [n]}$, where $x_i \in \cX$ and $y_i \in \cY$. Usually $\cX \subset \RR^d$, in which case a sample is simply a $d$-dimensional vector, whereas $\cY$ depends on the task at hand. For a regression task one usually takes $\cY = \RR$, whilst for a classification task on $C$ classes one usually assumes $\cY = \{1,2,\ldots,C\}$. We restrict ourselves to classification problems.

Supervised ML aims at \textit{learning} a map $h: \cX \rightarrow \cY$ that generalizes to unseen samples based on the provided labeled dataset $\cD$. 
A common assumption used to learn the mapping is that all data points in $\cD$ are sampled identically and independently (i.i.d.) from an unknown distribution $p(X,Y)$, where $X,Y$ are random variables taking values in $\cX$ and $\cY$, respectively. For a single realisation $(x, y)$, we abbreviate $p(x,y) = p(X\!=\!x,\!Y\!=\!y)$. 

The goal is to choose $h(\cdot) \in \cH$, 
where $\cH$ represents the hypothesis space, that minimizes the expected risk with respect to the underlying probability distribution~\cite{shalev2014understanding}. In other words, one wants to construct $h^*$ such that
\begin{equation}
\label{eq:expected_risk}
h^* = \arg \min_{h \in \cH} \mathbb{E}_{X,Y} \left( L(h(x), y) \right) = \arg \min_{h \in \cH} \int_\cX \int_\cY L(h(x), y) p(x, y)\,dy\,dx ,
\end{equation}
with $L(\hat{y}, y)$ being a loss function that penalizes wrongly predicted labels  $\hat{y}$. For example, $L(\hat{y}, y) = \mathbf{1} (\hat{y} = y)$ represents the 0-1 loss, commonly chosen for classification problems. 
Finding the optimal mapping $h^*$ is not feasible in practice: (1) the underlying probability $p(X,Y)$ is typically unknown and it can only be approximated using a finite number of samples, (2) even if the distribution were known, calculating the integral is intractable for many possible choices of $p(X,Y)$. Therefore, in practice one performs an empirical risk minimization (ERM) by solving 
$\hat{h} = \arg \min_{h \in \cH} \frac{1}{n} \sum_{i=1}^n L(h(x_i), y_i).$
Despite the fact that the model is learned using a finite number of data samples, the ultimate goal is to learn a model which generalizes to any sample originating from the underlying probability distribution, by approximating its posterior $p(Y|X)$.
Using $\hat{h}$ to approximate $h^*$ can run into what-is-known as ``overfitting'' to the training set $\cD$, which reduces the generalization property of the mapping.
However, advances in statistical learning theory managed to considerably lower the expected risk for many real-world applications whilst avoiding overfitting~\cite{friedman2001elements, vapnik2015uniform, shalev2014understanding}.  Altogether, any aspect of data quality for ML application development should not only be treated with respect to the dataset $\cD$ or individual data points therein, but also \emph{with respect to the underlying probability distribution the dataset $\cD$ is sampled from}.

\paragraph{Validation and Test}

Standard ML cookbooks suggest that the data should be represented by three disjoint sets to \emph{train}, \emph{validate}, and \emph{test}. 
The validation set accuracy is typically used to choose the best possible set of hyper-parameters used by the model trained on the training set.
The final accuracy and generalization properties are then evaluated on the test set. Following this, we use the term \textit{validation} for evaluating models in the pre-training phase, and the term \textit{testing} for evaluating models in the post-training phase.

\paragraph{Bayes Error Rate}
Given a probability distribution $p(X,Y)$, the lowest possible error rate achievable by \textit{any} classifier is known in the literature as the Bayes Error Rate (BER). It can be written as
\begin{equation}
\label{eq:bayes_error}
R_{X,Y}^* = \mathbb{E}_{X} \left[ 1- \max_{y\in\cY} p(y| x) \right],
\end{equation}
and the map $h_{opt}(x)=\arg \max_{y\in\cY} p(y \vert x)$ is called the \textit{Bayes Optimal Classifier}. It is important to note that, even though $h_{opt}$ is the best possible classifier (that is often intractable for the reason stated above), its expected risk might still be greater than zero, which results in the accuracy being at most $1 - R_{X,Y}^*$. In Section~\ref{sec:snoopy}, we will outline multiple reasons and provide examples for a non-zero BER.

\paragraph{Concept Shift}

The general idea of ML described so far assumes that the probability distribution $P(X, Y)$ remains fixed over time, which is sometimes not the case in practice~\cite{widmer1996learning, tsymbal2004problem, gama2014survey}. Any change of distribution over time is known as a \textit{concept shift}.
Furthermore, it is often assumed that both the feature space $\cX$ and label space $\cY$ remain identical over a change of distribution, which could also be false in practice. A change of $\cX$ or $p(X)$ (marginalized over $Y$) is often referred to as a \textit{data drift}, which can result in missing values for training or evaluating a model. We will cover this specific aspect in Section~\ref{sec:cpclean}. When a change in $p(X)$ modifies $p(Y\vert X)$, this is known as a \textit{real drift} or a \textit{model drift}, whereas when $p(Y\vert X)$ stays intact it is a \textit{virtual drift}. Fortunately, virtual drifts have little to no impact on the trained ML model, assuming one managed to successfully approximate the posterior probability distribution over the entire feature space $\cX$.

\section{MLOps Task 1: Effective ML Quality Optimization}
\label{sec:cpclean}

\begin{figure}[t!]
\centering
\includegraphics[width=0.9\textwidth]{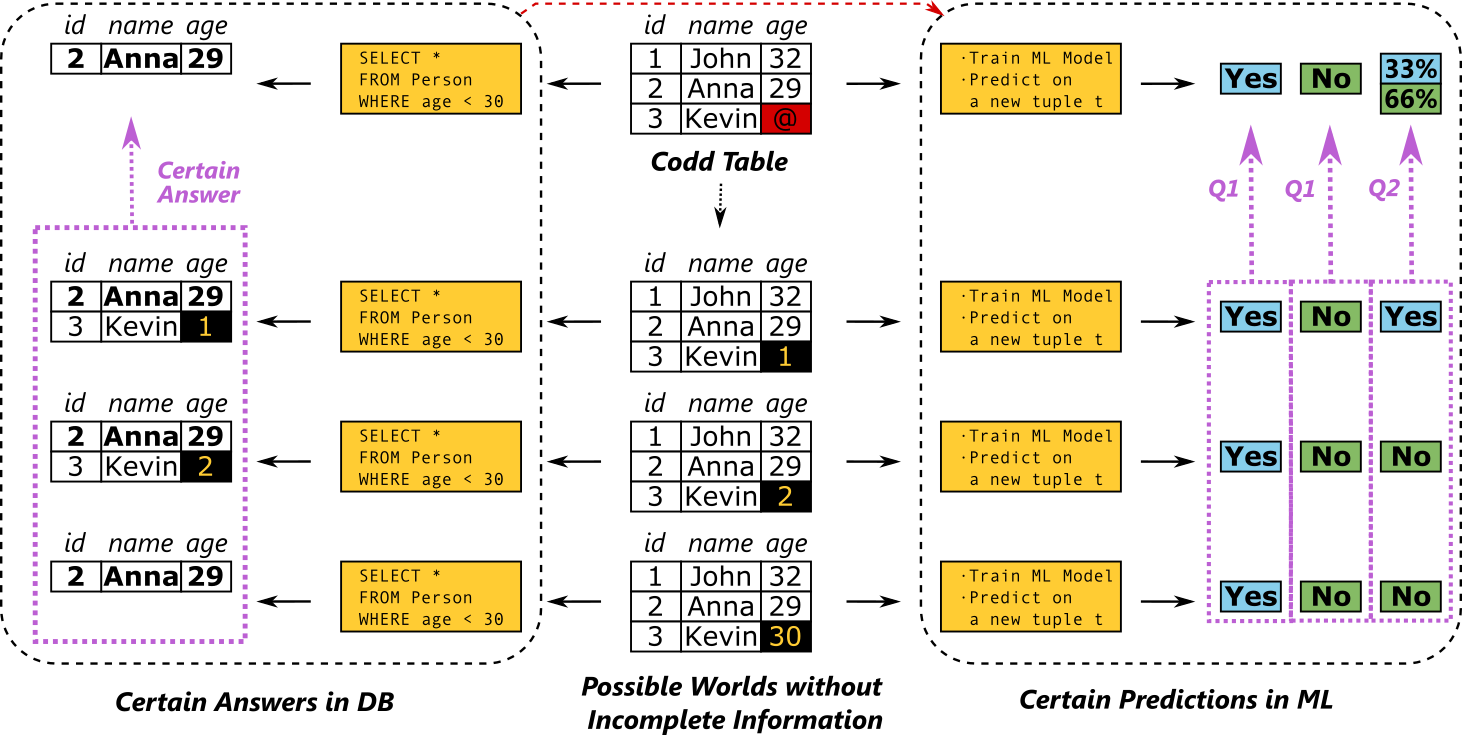}
\caption{Illustration of the relation between Certain Answers and Certain Predictions~\cite{karlavs2020nearest}. On the right, Q1 represents a \textit{checking query}, whereas Q2 is a \textit{counting query}.}
\label{fig:cpclean}
\end{figure}

\vspace{-0.5em}
One key operation in MLOps is seeking a way to 
improve the quality (e.g., accuracy) of a model.
Apart from trying new architectures and models,
improving the quality and quantity of the training
data has been known to be at least as important~\cite{li2019cleanml,fernandez2014we}.
Among many other approaches, data cleaning~\cite{ilyas2019data}, the practice of fixing 
or removing noisy and dirty samples,
has been a well-known strategy for improving 
the quality of data.

\vspace{-0.5em}
\paragraph*{MLOps Challenge}
When it comes to MLOps, a challenge is that 
not all noisy or dirty samples matter equally 
to the quality of the final ML model. In other words --
when ``propagating'' through the ML training process,
noise and uncertainty of different input samples
might have vastly different effects. As a result,
simply cleaning the input data artifacts either
randomly or agnostic to the ML 
training process might lead to a sub-optimal improvement
of the downstream ML model~\cite{li2019cleanml}.
Since the cleaning task itself is often 
performed ``semi-automatically'' 
by human annotators, with guidance from automatic
tools, the goal of a \textit{successful} cleaning strategy from an MLOps perspective should be to minimize the amount of human effort. This typically leads to a partially cleaned dataset, with the property that cleaning additional training samples would not affect the outcome of the trained model (i.e., the predictions and accuracy on a validation set are maintained).

 \vspace{-0.5em}
\paragraph*{A Data Quality View}
A principled solution to the above 
challenge requires a \textit{joint}
analysis of the impact of incomplete and noisy data in the training
set on the quality of an ML model trained over
such a set. Multiple seminal works have studied this problem, e.g., ActiveClean~\cite{krishnan2016activeclean}.
Inspired by these, we introduced 
a principled framework called CPClean that models
and analyzes such a 
noise propagation process together with principled cleaning algorithms based on 
sequential information maximization~\cite{karlavs2020nearest}.

\paragraph{Our Approach: Cleaning with CPClean}
CPClean directly models the noise propagation ---
the noises and incompleteness introduce
multiple possible datasets, called \emph{possible worlds} in relational database theory,
and the impact of these noises to 
final ML training is simply the \textit{entropy} 
of training multiple ML models, one for each of these possible worlds. Intuitively,
the smaller the entropy, the less impactful
the input noise is to the downstream 
ML training process.
Following this, we start by initiating all possible worlds (i.e., possible versions of the training data after cleaning) by applying multiple well-established cleaning rules and algorithms independently over missing feature values.
CPClean then operates in multiple iterations. At each round, the framework suggests the training data to clean that minimizes the \textit{conditional entropy} of possible worlds over the partially clean dataset. Once a training data sample is cleaned, it is replaced by its cleaned-up version in all possible worlds. At its core, it uses a \emph{sequential information-maximization} algorithm that finds an approximate solution (to this NP-Hard problem) with theoretical guarantees~\cite{karlavs2020nearest}.
Calculating such an entropy is often difficult,
whereas in CPClean we provide efficient algorithms which can calculate this term in polynomial time for a specific family of 
classifiers, namely k-nearest-neighbour
classifiers (kNN).

This notion of learning over incomplete data using \emph{certain predictions} is inspired by research on \emph{certain answers} over incomplete data~\cite{abiteboul1995foundations, suciu2011probabilistic, arenas1999consistent}. In a nutshell, the latter reasons about \emph{certainty} or \emph{consistency} of the answer to a given input, which consists of a query and an incomplete dataset, by enumerating the results over all possible worlds. Extending this view of data incompleteness to non-relational operator (e.g., an ML model) is a natural yet non-trivial endeavor, and Figure~\ref{fig:cpclean} illustrates the connection.

\paragraph{Limitations}
Taking the downstream ML model into account for prioritizing human cleaning effort is not new. ActiveClean~\cite{krishnan2016activeclean} suggests to use information about the \textit{gradient} of a fixed model to solve this task. Alternatively, our framework relies on consistent predictions and, thus, works on an unlabeled validation set
and on ML models that are not differentiable. 
In~\cite{karlavs2020nearest}
we use kNN as a proxy to an arbitrary classifier, given its efficient implementation despite exponentially many possible worlds. However, it still remains to be seen
how to extend this principled framework 
to other types of classifiers. Moreover, combining both approaches and supporting a labor-efficient cleaning approach for general ML models remains an open research problem.

\section{MLOps Task 2: Preventing Unrealistic Expectations}
\label{sec:snoopy}

In DevOps practices, new projects are typically initiated with a \textit{feasibility study}, in order to evaluate and understand the probability of success. The goal of such a study is to prevent users with unrealistic expectations from spending a lot of of money and time on developing solutions that are doomed to fail. However, when it
comes to MLOps practices, such a 
feasibility study step is largely missing --- we often
see users with high expectations, but with a very noisy dataset, starting an
expensive training process which is almost surely
doomed to fail.

\paragraph*{MLOps Challenge}
One principled way to model the feasibility
study problem for ML is to ask:
\textit{Given an ML task, defined by its training and validation sets, how to estimate the 
error that the best
possible ML model can achieve, without
running expensive ML training?}
The answer to this question 
is linked to a traditional ML problem, i.e.,
to estimate the \emph{Bayes error rate} (also called \emph{irreducible error}). It is a quantity related to the underlying data distribution and estimating it using finite amount of data is known to be a notoriously hard problem. Despite decades of study~\cite{cover1967nearest, fukunaga1987bayes, sekeh2018multi},
providing a practical BER estimator is still 
an open research problem and there are no known practical 
systems that can work on real-world large-scale datasets. One key challenge to make feasibility
study a practical MLOps step is to understand 
how to utilize decades of theoretical studies
on the BER estimation and which  
compromises and optimizations to perform.

\begin{figure}[t!]
\begin{subfigure}[t]{.32\textwidth}
\centering\captionsetup{width=.95\linewidth}
\includegraphics[width=0.95\linewidth]{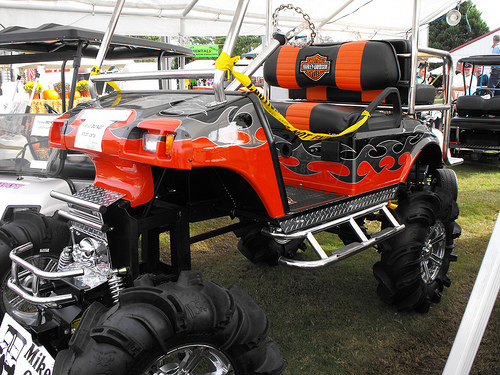}  
\caption{Image (\#6874) illustrating a non-unique label probability.}
\label{fig:imnet_missingfeature}
\end{subfigure}
\hfill
\begin{subfigure}[t]{.32\textwidth}
\centering\captionsetup{width=.95\linewidth}
\includegraphics[width=0.95\linewidth]{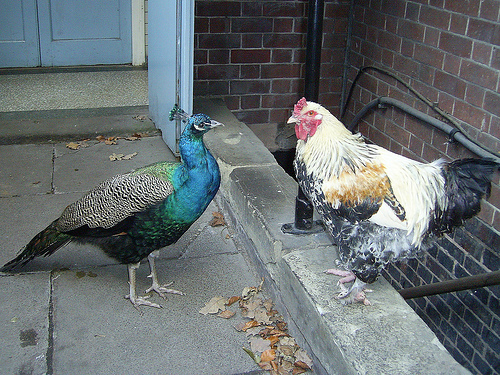}  
\caption{Image (\#4463) showing multiple classes for a fixed sample.}
\label{fig:imnet_missinglabel}
\end{subfigure}
\hfill
\begin{subfigure}[t]{.32\textwidth}
\centering\captionsetup{width=.95\linewidth}
\includegraphics[width=0.95\linewidth]{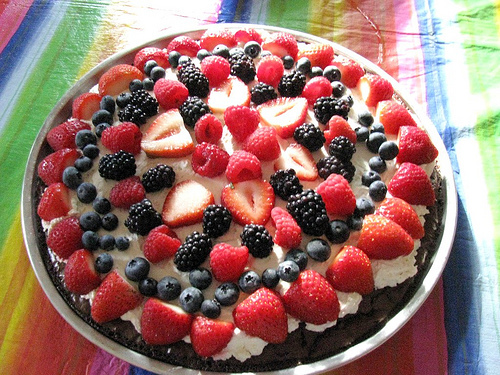}  
\caption{Image (\#32040) mislabeled as a ``pizza''.}
\label{fig:imnet_labelnoise}
\end{subfigure}
\caption{ImageNet examples from the validation set illustrating possible reasons for a non-zero Bayes Error.}
\label{fig:imagenet_examples}
\end{figure}

\subparagraph{Non-Zero Bayes Error and Data Quality Issues}
At the first glance, even understanding why the BER is not zero for every task can be quite mysterious --- 
\textit{if we have enough amount of data and 
a powerful ML model, what would stop us from 
achieving perfect accuracy?}
The answer to this is deeply connected to 
data quality.
There are two classical data quality dimensions that constitute the reasons for a non-zero BER: (1)~\textit{completeness} of the data, violated by an insufficient definition of either the feature space or label space, and (2)~\textit{accuracy} of the data, mirrored in the amount of noisy labels. 
On a purely mathematical level, the reason for a non-zero BER lies in \textit{overlapping} posterior probabilities for different classes, given a realisable input feature.
More intuitively, for a given sample the label might not be unique.
In Figure~\ref{fig:imagenet_examples} we illustrate some real-world examples from the validation set of \textit{ImageNet}~\cite{deng2009imagenet}.
For instance, Figure~\ref{fig:imnet_missingfeature} is labeled as a golfcart (n03445924) 
whereas there is a non-zero probability that the vehicle belongs to another category, for instance a tractor (n04465501) -- additional features can resolve such an issue by providing more information and thus leading to a single possible label.
Alternatively, there might in fact be multiple ``true'' labels for a given image. Figure~\ref{fig:imnet_missinglabel} shows such an example, where the posterior of class rooster (n01514668) is equal to the posterior of the class peacock (n01806143), despite being only labeled as a rooster in the dataset -- changing the task to a multi-label problem would resolve this issue.
Finally, having noisy labels in the validation set yields another sufficient condition for a non-zero BER.
Figure~\ref{fig:imnet_labelnoise} shows such an example, where a pie is incorrectly labeled as a pizza (n07873807).

\paragraph*{A Data Quality View}
There are two main challenges in building 
a practical BER estimator for ML models to
characterize the impact of data quality to 
downstream ML models: (1) the computational requirements and (2) the choice of hyper-parameters. Having to estimate the BER in today's high-dimensional feature spaces
requires a large amount of data in order to give a reasonable estimate in terms of accuracy, which results in a high computational cost. Furthermore, any practical estimator should be insensitive to different hyper-parameters, as no information about the data or its underlying distribution is known \emph{prior to} running the feasibility study.

\begin{wrapfigure}{r}{0.5\textwidth}
\vspace{-2em}
\centering
\includegraphics[width=0.5\textwidth]{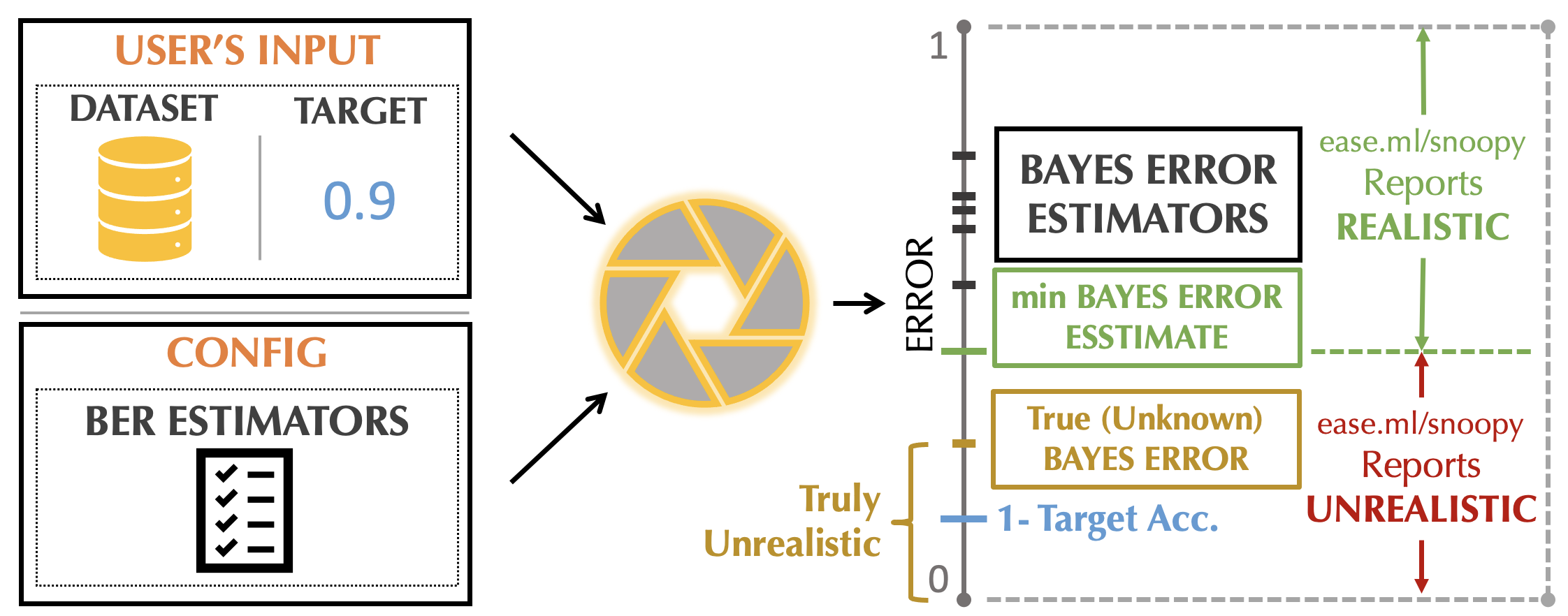}
\label{fig:snoopy}
\vspace{-2em}
\end{wrapfigure}

\paragraph{Our Approach: \texttt{ease.ml/snoopy}}
We design a novel BER estimation method that (1) has no hyper-parameters to tune, as it is based on nearest-neighbor estimators, which are non-parametric; and (2) uses pre-trained embeddings, from public sources such as PyTorch Hub or Tensorflow Hub\footnote{\url{https://pytorch.org/hub} and \url{https://tfhub.dev}}, to considerably decrease the dimension of the feature space. The aforementioned functionality of performing a feasibility study using \texttt{ease.ml/snoopy} is illustrated in the above figure. For more details we refer interested readers to both the full paper~\cite{renggli2020automatic} and the demo paper for this component~\cite{renggli2020ease}. The usefulness and practicality of this novel approach is evaluated on well-studied standard ML benchmarks through a new evaluation methodology that injects label noise of various amounts and follows the evolution of the BER~\cite{renggli2020automatic}. It relies on our theoretical work~\cite{rimanic2020convergence}, in which we furthermore provide an in-depth explanation for the behavior of kNN over (possibly pre-trained) feature transformations by showing a clear trade-off between the increase of the BER and the boost in convergence speed that a transformation can yield.

\paragraph{Limitations}
The standard definition of the BER assumes that both the training and validation data are drawn i.i.d. from the \textit{same} distribution, an assumption that does not always hold in practice.
Extending our work to a setup that takes into account two different distributions for training and validation data, for instance as a direct consequence of applying data programming or weak supervision techniques~\cite{ratner2017snorkel}, offers an interesting line of future research, together with developing even more practical 
BER estimators for the i.i.d. case.

\section{MLOps Task 3: Rigorous Model Testing Against Overfitting}
\label{sec:ci}

One of the major advances in running fast and robust cycles in the software development process is known as continuous integration (CI)~\cite{duvall2007continuous}. The core idea is to carefully define and run a set of conditions in the form of tests that the software needs to successfully pass every time prior to being pushed into production. This ensures the robustness of the system and prevents unexpected failures of production code even when being updated.
However, when it comes to MLOps, the traditional 
way of reusing the same test cases repeatedly can 
introduce serious risk of overfitting, thus 
compromise the test result.

\paragraph*{MLOps Challenge}
In order to generalize to the unknown underlying probability distribution, when training an ML model, one has to be careful not to overfit to the (finite) training dataset. However, much less attention has been devoted to the statistical generalization properties of the \textit{test set}.
Following best ML practices, the ultimate testing phase of a new ML model should either be executed only once per test set, or has to be completely obfuscated from the developer. Handling the test set in one way or the other ensures that no information of the test set is \emph{leaked} to the developer, hence preventing potential overfitting. Unfortunately, in ML development environments it is often impractical to implement either of these two approaches.

\paragraph*{A Data Quality View}
Adopting the idea of continuously testing and integrating ML models in productions has two major caveats: (1) test results are inherently random, due to the nature of ML tasks and models, and (2) revealing the outcome of a test to the developer could mislead them into overfitting towards the test set.
The first aspect can be tackled by using well-established concentration bounds known from the theory of statistics.
To deal with the second aspect, which we refer to as the \textit{timeliness} property of testing data, there is an approach pioneered by Ladder~\cite{blum2015ladder}, together with the general area of \emph{adaptive analytics} (cf.~\cite{dwork2015reusable}), that enable multiple reuses of the same test set with feedback to the developers. The key insight of this line of work is that the \textit{statistical power} of a fixed dataset shrinks when increasing the number of times it is reused. In other words, requiring a minimum statistically-sound confidence in the generalization properties of a finite dataset limits the number of times that it can be reused in practice.

\paragraph{Our Approach: Continuous Integration of ML Models with \texttt{ease.ml/ci}}

\begin{figure}
\centering
\includegraphics[width=1.0\textwidth]{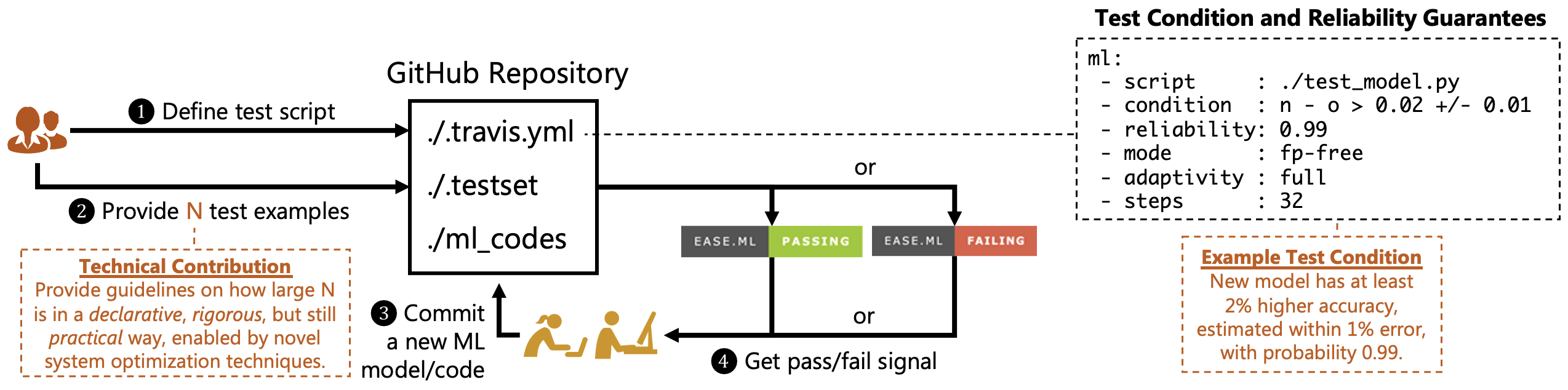}
\caption{The workflow of \texttt{ease.ml/ci}, our CI/CD engine for ML models~\cite{renggli2019continuous}.}
\label{fig:ci}
\end{figure}

As part of the \texttt{ease.ml} pipeline, we designed a CI engine to address both aforementioned challenges. The workflow of the system is summarized in Figure~\ref{fig:ci}. The key ingredients of our system lie in (a) the syntax and semantics of the test conditions and how to accurately evaluate them, and (b) an optimized \emph{sample-size estimator} that yields a budget of test set re-uses before it needs to be refreshed. For a full description of the workflow as well as advanced system optimizations deployed in our engine, we refer the reader to our initial paper~\cite{renggli2019continuous} and the followup work~\cite{karlavs2020building}, which further discusses the integration into existing software development ecosystems.

We next outline the key technical details falling under the 
general area of ensuring generalization properties of finite data used to test the accuracy of a trained ML model repetitively.

\subparagraph{Test Condition Specifications}

A key difference between classical CI test conditions and testing ML models lies in the fact that the test outcome of any CI for ML engine is inherently probabilistic. Therefore, when evaluating result of a test condition that we call a \textit{score}, one has to define the desired confidence level and tolerance as an $(\epsilon,\delta)$ requirement. Here $\epsilon$ (e.g., $1\%$) indicates the size of the confidence interval in which the estimated score has to lie with probability at least $1-\delta$ (e.g., $99\%$). For instance, the condition \verb|n - o > 0.02 +/- 0.01| requires that the new model is at least 2 points better in accuracy than the old one, with a confidence interval of 1 point. Our system additionally supports the variable \verb|d| that captures the fraction of different predictions between the new and old model.  
For testing whether a test condition passes or fails one needs to distinguish two scenarios. On one hand, if the score lies outside the confidence interval (e.g., \verb|n - o > 0.03| or \verb|n - o < 0.01|), the test immediately passes or fails.
On the other hand, the outcome is ill-defined if the score lies inside the confidence interval. Depending on the task, user can choose to allow \emph{false positive} or \emph{false negative} results (also known as ``type I'' and ``type II'' errors 
in statistical hypothesis testing), after which all the scores lying inside the confidence interval will be automatically rejected or accepted.

\subparagraph{Test Set Re-Uses}

In the case of a non-adaptive scenario in which no information is revealed to the developer after running the test, the least amount of samples needed to perform $H$ evaluations with the same dataset is the same as running a single evaluation with $\delta / H$ error probability, since the $H$ models are independent.
Therefore, revealing any kind of information to the developer would result in a dataset of size $H$ multiplied by the number of samples required for one evaluation with $\delta / H$ error. However, this trivial strategy is very costly and usually impractical. The general design of our system offers a different approach that significantly reduces the amount of test samples needed for using the same test set multiple times. More precisely, after every commit the system only reveals a binary pass/fail signal to the developer. Therefore, there are $2^H$ different possible sequences of pass/fail responses, 
which yields that the number of samples needed for $H$ iterations is the same as running a single iteration with $\delta / 2^H$ error probability -- much smaller than the previous $\delta / H$ one. 
We remark that further optimizations can be deployed by making use of the structure or low variance properties that are present in certain test conditions, for which we refer the interested readers to the full paper~\cite{renggli2019continuous}.

\paragraph{Limitations}

The main limitation consists of the worst-case analysis which happens when the developer acts as an adversarial player that aims to overfit towards the hidden test set. Pursuing other, less pragmatic approaches to model the behavior of developers could enable further optimization to reduce the number of test samples needed in this case.
A second limitation lies in the lack of ability to handle concept shifts. Monitoring a concept shift could be thought of as a similar process of CI -- instead of fixing the test set and testing multiple models, one could fix a single model and test its generalization over multiple test sets. From that perspective, we hope that some of the optimizations that we have derived in our work could potentially be applied to monitoring concept shifts as well. Nevertheless, this needs further study and forms an interesting research path for the future.

\newpage
\section{MLOps Task 4: Efficient Continuous Quality Testing}
\label{sec:model_picker}

One of the key motivations for DevOps principles in the first place is the ability to perform fast cycles and continuously ensure the robustness of a system by quickly adapting to changes. At the same time, both are well-known requirements from traditional software development that naturally extend to the MLOps world.
One challenge faced by many MLOps practitioners is the necessity
to deal with the shift of data distributions when models are
in production. When new production 
data comes from a different (unknown) distribution,
models trained over previously seen 
data distributions might not perform well
anymore. 

\paragraph*{MLOps Challenge}
While there has been various research on 
automatic domain adaption~\cite{shimodaira2000improving, sugiyama2008direct, zhang2013domain}, we identify a 
different challenge when presented with a collection of models, each of which could be a ``staled model'' or 
an automatically adapted model given some domain adaption method. 
This scenario is quite common in many 
companies --- they often train distinct models on
different slices of data independently (for instance one model for each season) and 
automatically adapt each of these
models using different methods for new data. As a result,
they often have access to a large 
set of models that could be deployed, hoping to know which one to use given
a fresh collection of production data (e.g., the current time period such as the current day).
The challenge is, given an unlabeled 
production data stream, to pick the model that
performs best. From the MLOps perspective,
the goal is to minimize the amount of labels needed to 
acquire in order to make such a distinction.

\paragraph*{A Data Quality View}
Concept shift is by its definition related to the \textit{timeliness} properties of the data. The available pre-trained models are intended to capture the changes of training data over time. Naturally, simple rules can be applied to choose the current model if one has access to some meta information about both the current timestamp and the pre-trained models (e.g., the current weekday and for each model the day it represents).
This problem gets particularly difficult when there are no such meta-data available. In that case, having access to a fully labeled clean dataset would result in trivially selecting the pre-trained model that achieves the highest accuracy. Collecting labels for a large enough test set is very costly in practice compared to simply gathering a set of unlabeled samples though. The reason is that accurately labeling samples requires human, if not expert level, annotators. Consequently, one wishes to robustly solve the problem of \textit{picking} the best model for the current time span with the fewest amount of labeling effort necessary and thus relying on \textit{incomplete} test data with respect to their labels.

\paragraph{Our Approach: \texttt{ease.ml/ModelPicker}}
Model Picker is an online model selection approach to selectively sample instances that are informative for ranking pre-trained models~\cite{karimi2020online}. Specifically, given a set of pre-trained models and a stream of unlabeled data samples that arrive sequentially from a data source, the Model Picker algorithm answers when to query the label of an instance, in order to pick the best model under limited labeling budget. We conduct a rigorous theoretical analysis to show that Model Picker has no regret for adversarial streams (e.g., non-i.i.d. data), and is effective in online prediction tasks for both adversarial and stochastic streams. Moreover, our theoretical bounds match (up to constants) those of existing online algorithms that have access to all the labels. 

\paragraph{Limitations}
One immediate extension of Model Picker is towards a setting in which the user at once has access to a pool of unlabeled data samples. In such a \textit{pool-based sampling} case~\cite{settles2009active}, one can rank the entire collection of data samples to select the most informative example instead of scanning through the data sequentially to decide whether to query a label or not. Despite the applicability of Model Picker to such a scenario where one can form a stream by sampling i.i.d. from the pool of samples, the availability of entire data collection can be exploited to further reduce the annotation costs with a more specialized strategy for pool-based scenarios.

\newpage
\section{Moving Forward}
\label{sec:limitations}

We have briefly described four 
of our previous works with a unified theme ---
all of them provide, in our opinion, \textit{
functionalities that are useful to 
facilitate a better MLOps process}, which, on 
the flip side, introduce new fundamental 
technical problems that require us to 
\textit{jointly analyze the impact
of data quality issues to downstream ML processes.}
When studying these technical problems,
we often need to 
go beyond an ML-agnostic view 
of data quality and, instead, need to develop 
new methods that \emph{simultaneously} combine the two aspects of ML \emph{and} data quality.
Despite the progress that we have made so far, this endeavor is still at its early stages. 
In the following, we present two future directions that, in our opinion,
are necessary to facilitate both MLOps as an important
functionality and ML-aware data quality as a fundamental research area.

\paragraph*{ML-Aware Data Quality}
From a technical perspective,
jointly understanding data quality and 
downstream ML processes is both interesting
and challenging. All results we discussed
in this paper are arguably limited~\cite{karlavs2020nearest, renggli2020automatic, karimi2020online, renggli2019continuous} --- after starting 
from a principled formulation of a 
problem, reaching fundamental 
computational challenges within these  
principled frameworks is inevitable. We get around
those by either (1) opting for simpler 
proxy models for which we can derive 
stronger results and/or more efficient algorithms
(e.g., kNN used in \texttt{ease.ml/snoopy}~\cite{renggli2020automatic}
and CPClean~\cite{karlavs2020nearest})
or (2) optimizing for specific cases 
commonly used in practice (e.g., the patterns in \texttt{ease.ml/ci}~\cite{renggli2019continuous} that we optimized for).
To further facilitate MLOps in general,
we are in dire need for an 
ML-aware data quality that is not only principled,
but also practical for a larger 
collection of scenarios and ML models.
These are all technically challenging --- simply
extending the methods that we developed is unlikely to succeed.
We hope that our current endeavors~\cite{karlavs2020nearest, renggli2020automatic, karimi2020online, renggli2019continuous} can serve, in
some ways, as ``examples of failures''
that other researchers can draw inspirations from.

\paragraph{Beyond Accuracy}
Another common limitation of our work~\cite{karlavs2020nearest, renggli2020automatic, karimi2020online, renggli2019continuous}
is that they all focus on improving the \textit{accuracy} of an ML model artifact. Although this is one of the most important 
aspects of model quality, recently researchers have
also identified multiple interesting dimensions
of model quality such as robustness, fairness,
and explainability. 
Even though we expect these quality dimensions to
become the core of the MLOps process in the future,
how to extend functionalities that we 
developed for improving accuracy to 
these quality dimensions is still an open question.
Jointly analyzing the impact of all data-quality dimensions with respect to more than a single metric that quantifies ML models is a large and promising research area that we believe will provide further understanding and improvements of the MLOps process.

\section*{Acknowledgments}
\small
We are grateful to many collaborators that we have 
been working with over the years, especially (in the context of this paper) those
who also contributed to the development of techniques in \cite{karlavs2020nearest, renggli2020automatic, karimi2020online, renggli2019continuous} including Peng Li, Prof. Xu Chu, Mohammad Reza Karimi, and Prof. Andreas Krause. These technical results would not be possible without their contributions.

CZ and the DS3Lab gratefully acknowledge the support from the Swiss National Science Foundation (Project Number 200021\_184628), Innosuisse/SNF BRIDGE Discovery (Project Number 40B2-0\_187132), European Union Horizon 2020 Research and Innovation Programme (DAPHNE, 957407), Botnar Research Centre for Child Health, Swiss Data Science Center, Alibaba, Cisco, eBay, Google Focused Research Awards, Oracle Labs, Swisscom, Zurich Insurance, Chinese Scholarship Council, and the Department of Computer Science at ETH Zurich.

\newpage
\scriptsize
\bibliographystyle{abbrv}
\bibliography{ieeedeb}

\end{document}